# Vision-Based Automatic Groceries Tracking System - Smart Homes

Divya Mereddy

Computer Science

Vanderbilt University

Nashville, TN

divya.mereddy@vanderbilt.edu

*Abstract— With advanced AI, while every industry is growing at rocket speed, the smart home industry has not reached the next-generation. There is still a huge leap of innovation that needs to happen before we call a home a 'Smart home'. A Smart home should predict residents' needs and fulfill them in a timely manner. One of the important tasks of maintaining a home is timely grocery tracking and supply maintenance. Grocery tracking models are very famous in the retail industry but they are nonexistent in the common household. Groceries detection in household refrigerators or storage closets is very complicated compared to retail shelving data. In this paper, home grocery tracking problem is resolved by combining retail shelving data and fruits dataset with real-time 360 view data points collected from home groceries storage. By integrating this vision-based object detection system along with supply chain and user food interest prediction systems, complete automation of groceries ordering can be achieved.*

## I. INTRODUCTION (HEADING 1)

In recent years, there has been a huge growth in the world of intelligent devices for home automation. Such gadgets are designed in order to ease the interaction between people and daily home duties. As both AI and smart home technologies advance, more use cases emerge. It seems as though more possibilities present themselves for home automation, so it's exciting to see how they can make life easier for homeowners using applications like smart closets, grocery tracking, food suggestions etc. With lack of drastic advancement in smart home AI systems, we still see human efforts that could be spent elsewhere. Automating such repetitive tasks might free up more time and give out more efficient results.

Today, we can find that the majority of similar tasks are already automated in AI industry. But we do not see a collaborated automation of such tasks in household activities implemented effectively. Groceries tracking systems are already implemented in retail stores. Similar implementations can be introduced in household activities and modified to suit household usage. The major challenge with this implementation is object detection for complicated refrigerator.

Store groceries shelving tracking is well organized. It's therefore easy to track though the amount of goods is high. When it comes to a groceries closet or refrigerator, it is very complicated because it's very unorganized, and small quantities can be stored in multiple locations within the storage space. Sometimes one fruit or piece of fruit can be stored in multiple places within refrigerator. Predicting the needs of the user is also a big problem but definitely predictable using a supply chain system[9], user inputs, and interest prediction systems[8]. A grocery tracking model is an AI asset that provides an enhanced level of automation without the need for human input. This can be done through a combination of sensors [1],[2], and vision[4], which are able to collect data based on the location, movement, and status of a given asset. This data is then fed into an AI system that uses predictive modeling to generate real-time updates of each asset quantity. In our paper, we are developing home grocery tracking system using computer vision.

## II. RESEARCH IN INDUSTRY

Overall the current research in this domain is still unsaturated, showing immense potential for future developments., particularly in vision-based applications, such as home grocery tracking. [1] has developed a similar project centered around prelabeled boxes and their respective weights.

### A. Smart grocery tracking systems based on sensors

A team from JEC Jabalpur M. P, India [1] has developed groceries track systems based on the weight of the box. But their functionalities are very limited. Different goods cannot be combined. We can find similar research from some other institutes [2][3].

### B. Refrigerator Industry

In the refrigerator industry, prominent companies such as Samsung, LG, and others have made commitments to introduce smart features in their products [5]. While some progress has been made in this direction, certain features still await automation, and specific aspects of research, particularly in groceries tracking, are yet to be initiated. As a result, achieving a fully automated groceries tracking system remains an elusive goal.

### C. Store Groceries Tracking

As mentioned previously, home groceries tracking is very similar to a store grocery tracking system. In-store grocery tracking domain, extensive research[4] is going on. But these techniques are not implemented in more complex home groceries system yet.

This project does not concentrate on the sensors technique. In our paper, we present an innovative approach utilizing the newly created refrigerator dataset[10] specially curated for this purpose: the implementation of YOLO for a smart home application. Although the task of grocery shelfing prediction for common households is more intricate than store groceries, we have successfully developed a unique model tailored to address this complexity.

III. BACKGROUND

Groceries tracking system is a part of smart home and it mainly integrates with food suggestions system and supply chain grocery ordering system. AI food suggestions system[8] provides automated food suggestions to users and integrates the user-selected food menu with the complete smart home system to predict the goods' needs. AI-based supply chain system[9] predicts the needs of users and integrate with grocery tracking systems. For this project, we developed a grocery tracking system for a basic weekly schedule of one person. It's a simple week plan with few repeated items as explained below.

As this problem was not researched before but again similar to a few industry-famous groceries tracking models, some of the existing data along with custom-built data was used. This project predicts the basic need of single user. It's an attempt to validate whether advanced AI systems are positioned to fulfill the needs of a human in simple user case.

Considering below as the list of the groceries user prefer to have in his refrigerator every day, in this project we will track them only.

TABLE I. TRACKING GROCERIES LIST

|   | *Groceries* |
|---|---|
| 1 | Banana |
| 2 | Avacado |
| 3 | Milk |
| 4 | Strawberries |
| 5 | Blueberries |
| 6 | tamatos |
| 7 | carrots |
| 8 | Salad Mix |
| 9 | egg white |

IV. DATA SOURCES

As this topic has not been researched before, new real-time data creation was needed. However, existing data sources are utilized to train our deep learning models, and real-time data created from the project environment is also utilized to customize the model predictions to this use case. Custom-created data has improved the efficiency of the model significantly.

*A. Needed data*

We need a large amount of image data of every single piece of grocery we can ever find in a refrigerator. The user case that we are considering in this paper is simple as shown in the Tracking Groceries list. We need minimum data that represent this list. To list them, we need fruits, vegetables, different brands of food products, milk, and many more. As explained above in our user case we need basic items like veggies, milk, water etc.

*B. Available data*

We can get the information from the groceries database. So we tried to utilize open-source grocery tracking data using Recognition of supermarket products using Computer Vision publication [2].

1. *Fruits 360 dataset[11]:* A dataset of images containing fruits and vegetables. It contains different types of fruits and vegetables (including different varieties) like Apples, Apricot, Avocado, Avocado ripe, Banana, and many more. But we considered only the items we are tracking shown in Table 1 Tracking Groceries List. It has a total Number of classes: 131 (fruits and vegetables) of image size: 100x100 pixels.

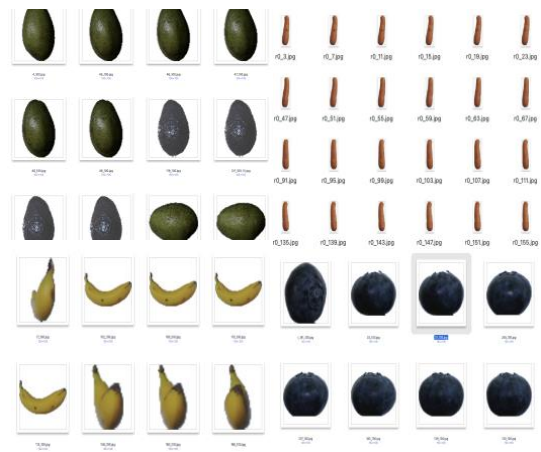
Fig. 1. Fruits 360 Dataset

2. *Groceries tracking data set [12] SKU110:*

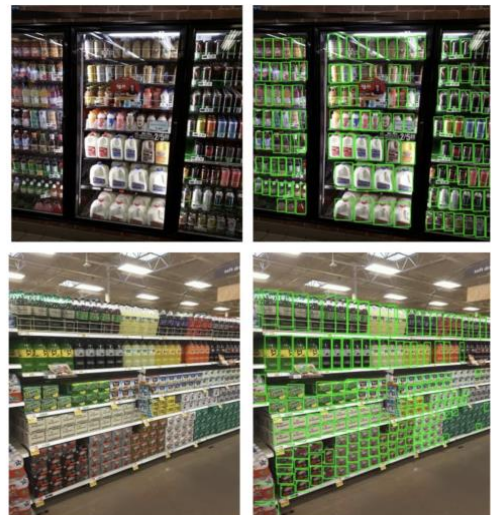
Fig. 2. Store tracking data - SKU110k dataset

SKU110k dataset is based on images of retail objects in a densely packed setting. It provides training, validation and test set images and the corresponding .csv files which contain information for bounding box locations of all objects in those images. The .csv files have object bounding box information written in the following columns: image_name,x1,y1,x2,y2, class,image_width,image_height where x1,y1 are top left coordinates of bounding box and x2,y2 are bottom right coordinates of bounding box, rest of parameters are self-explanatory.

An example of parameters of train_0.jpg image for one bounding box, is shown below. There are several bounding boxes for each image, one box for each object. We also aggregated [22] data set, consisting of the images of the shelf of the grocery store and annotations of the bounding boxes in the form of text files.

### C. Data sets needed to create (custom dataset)

Though the previously openly available data sets we considered are capturing fruits, vegetables, and groceries, they don't represent the home refrigerator/ groceries closet data. Home grocery storage system is complicated and messy. In addition, as the Fruits 360 dataset doesn't have object localization, the model was not able to predict multi outputs. The groceries dataset has all types of tracking items along with localization but, as discussed before, groceries data is very well organized compared to the home groceries dataset, which makes it difficult to predict accurately. To solve this problem, we needed to create real-time data to improve the performance of our model to predict the groceries in the refrigerator/closet accurately. To improve the dataset size and add additional capabilities to the model, like detecting objects from blurred veggies, we used data augmentation steps. We created more data using rotating, zooming techniques, and more as shown below.

**Preprocessing:**

Auto-Orient: Applied

Resize: Stretch to 640x640

**Augmentations:**

Outputs per training example: 3

Flip: Horizontal

Rotation: Between -15° and +15°

Grayscale: Apply to 25% of images

Hue: Between -25° and +25°

Saturation: Between -25% and +25%

Exposure: Between -25% and +25%

Blur: Up to 2.5px

Mosaic: Applied

Bounding Box: Shear: ±15° Horizontal, ±15° Vertical

Bounding Box: Exposure: Between -25% and +25%

Bounding Box: Noise: Up to 5% of pixels

This dataset is created from our research environment to customize the model predictions for this use case. In addition to the custom dataset created, we also included open-sourced refrigerator image data to capture refrigerator storage pictures based on the keywords method.

As mentioned [13] article, we also tried to create synthetic data based on a video that covers a 360 degrees view of a product. It helped us with the dataset size and helped to improve accuracy. We considered sub-data from all datasets mentioned above such that it can represent Table 1 Tracking Groceries List. we utilized Roboflow[14] for labeling unlabeled data. The final data dataset is this located at [10].

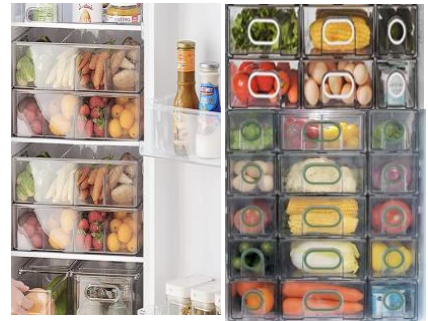

Fig. 3. Custom refrigerator/ closet groceries dataset

V. MODEL BUILDING

### A. Complexities & Challenges

While this problem is similar to store-based grocery tracking and fruit object detection, to achieve good accuracy we had to resolve the below complexities which are not present in either store-based grocery tracking or fruit object detection.

1. One of the main complexity of the project is that none of the openly available data represents the use case.
2. Capturing the complex common man refrigerator/ groceries closet data through images is difficult. As the common man tends to save his groceries in small amounts throughout their refrigerator (sometimes throughout their kitchen space).
3. The shelves are typically cluttered and often not organized in a regular fashion.
4. Ideal images of different products available to the vision system are often taken using different cameras resulting in different distributions of image intensities.
5. Also, due to different imaging parameters, the length of the product (in some units of length, say, cm) is mapped to different pixel resolutions for the product and shelf.

### B. Model Development

We can leverage vision models to gather information by employing micro cameras strategically placed inside the refrigerator at multiple locations. These cameras can capture images from various angles, allowing us to comprehensively assess the contents and overall condition of the refrigerator. Consequently, we can obtain a holistic view of the refrigerator's contents and generate an accurate grocery list based on the captured data. The utilization of multiple cameras, along with their ability to capture multi-angle pictures, has proven to be immensely valuable. This approach has greatly enhanced our capability to capture view of the refrigerator's contents. We developed two systems using ResNet[15] and YOLO[16].

*1. ResNet Model:*

For the initial training phase, we utilized the fruits360 dataset. The development of our model involved leveraging the power of ResNet networks. We have extended [17] fruit detection based on Fruits 360 dataset. As described this type of network makes use of convolutional layers, pooling layers, ReLU layers, fully connected layers, and loss layers. Utilizing the ResNet architecture, we successfully implemented fruit detection and classification, enabling us to assign relevant tags to each item. It was able to detect only single fruit(grocery item) at a time as shown in Fig 5. Because of the nature of the training data, the model was able to detect objects properly if there is only a single item in the complete input picture. To utilize this in real-time data we need to break down the whole storage into individual items and make predictions. Instead of this, we can also achieve similar results with multiple models or architecture changes to detect multi items. However, as the number of items to detect grew, the complexity of the system increases significantly. Instead of opting for these complicated methods, we went with an alternative approach by adopting an object-localized data-based multi-object detection model. This decision was made to overcome the challenges posed by a larger number of items, ensuring improved accuracy and efficiency in the detection and classification process.

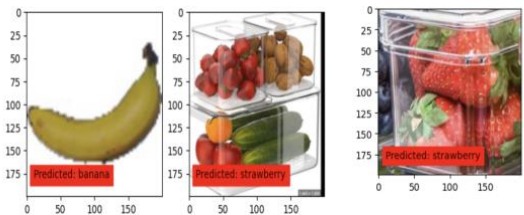

Fig. 4. ResoNet – Model Predictions

*2. Model Improvements - Object localization data with YOLO:*

Furthermore, it is important to note that the dataset used for the initial training of our model solely consisted of fruit 360 pictures. As a result, the model's performance was suboptimal when it came to predicting fruits from real-time refrigerator images, primarily due to the presence of multiple grocery items stacked together within the refrigerator environment.

We leveraged the power of the YOLO system to develop a custom-trained machine-learning solution capable of proficiently performing object localization and multi-object detection tasks. In addition to utilizing the Fruits 360 dataset and retail groceries datasets, we also incorporated openly available refrigerator groceries storage pictures data. Furthermore, we created additional data within the scope of this project to further enhance the customization of the YOLO model.

As discussed in the datasets section, the YOLO model was trained using a diverse range of data sources. These sources included retail groceries data, a fruit dataset, open-source data specific to refrigerator storage keywords, as well as custom-built data generated from real-time experiment environments. To facilitate the process of label creation and data augmentation, Roboflow was utilized. The inclusion of both the fruits dataset and retail shelf dataset provided the model with the ability to predict fruits and groceries effectively. However, to further enhance the model's understanding of the environment, we incorporated customized images, keyword-based images, and 360-view(based on [13]) pictures. This additional data allowed the model to gain a comprehensive 360-degree perspective, resulting in significant improvements(6%) in its predictive accuracy.

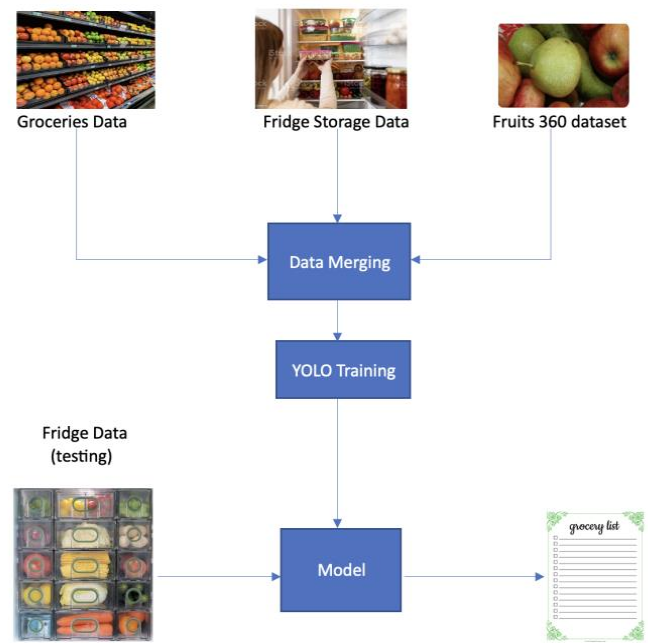

Fig. 5. Data Sources And Process Flow

The implemented model showcases both simplicity and innovation. YOLO models have become commonplace in the industry, particularly for groceries detection in the retail domain. However, the application of these detection models to home refrigerator data is a novel and viable approach. We encountered challenges in adapting the model to the complexities of real-time home refrigerator groceries storage, which differs significantly from retail shelving data. To overcome these challenges, we enhanced the model's performance by incorporating additional cameras and collecting more data for accurate predictions, regardless of the environment's complexity. shelves with storage boxes exhibit a relatively lower level of intricacy compared to shelves with unorganized items. Consequently, the shelves equipped with storage boxes demonstrated better performance metrics when compared to the data obtained from unorganized shelves.

VI. MODEL METRICS

The following 3 parameters are commonly used for object detection tasks: · GioU is the Generalized Intersection over Union which tells how close to the ground truth our bounding box is. Objectness shows the probability that an object exists in an image. Here it is used as a loss function. mAP is the mean Average Precision telling how correct are our bounding box predictions on average. It is area under the curve of precision-recall curve. It is seen that Generalized Intersection over Union (GIoU) loss and objectness loss decrease both for training and validation. Mean Average Precision (mAP) however is at 0.7 for bounding box IoU threshold of 0.5. Recall stands at 0.8 as shown below.

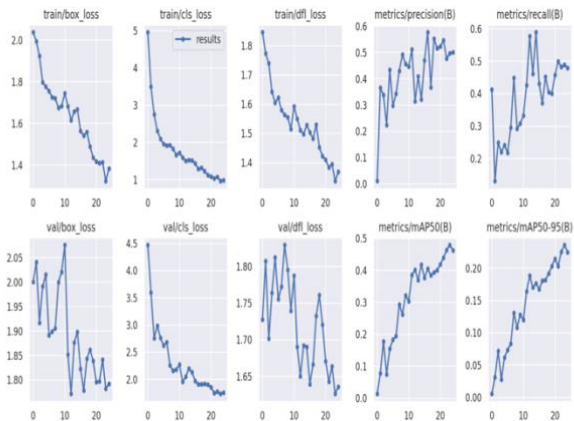

Fig. 6. Model Metrics

Predictions:

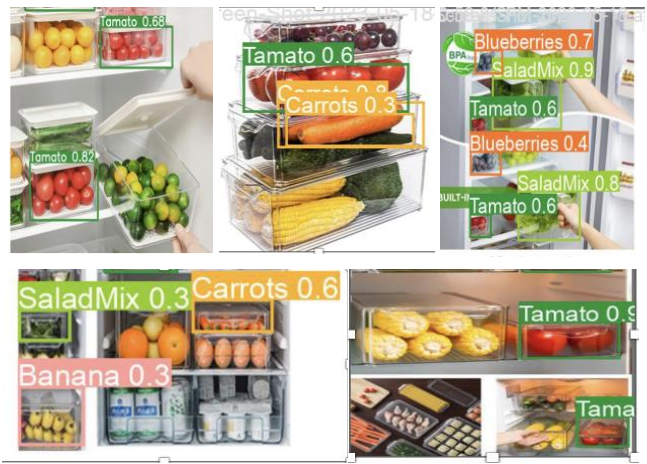

Fig. 7. Model Predictions

Based on the detected groceries and supply chain user needed groceries list, the model decides whether to create an order or not.

After using the application for a period, we observed a decline in performance. This decline can be attributed to the prolonged storage of groceries in boxes, resulting in the formation of a layer of water around the vegetables and inside the boxes as shown in fig 9. Consequently, the images of the groceries became blurry, causing confusion for the system. The system had been trained on higher-quality images, leading to a disparity between the trained picture quality and the real-time blurred picture data. Overall the performance was created by approximately the percentage of the moisture groceries. To address this issue, we took proactive measures by labeling new instances of "moisture" or "condensation inside food container" groceries and incorporating them into the training dataset. As we used a combination of different datasets for our training purpose, we utilized the concept of train-dev to ensure there is no data mismatch problem. In addition, our metrics are calculated using real-time data. Overall performance of the model in the case of "moisturized food or containers" was comparatively less.

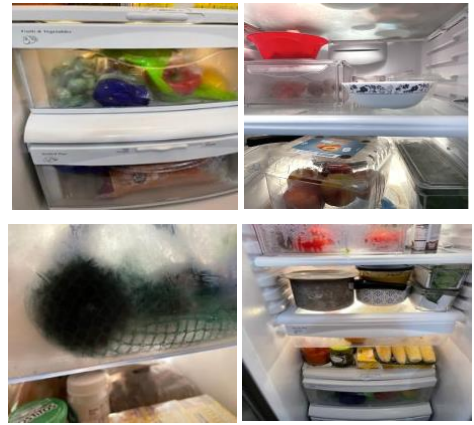

Fig. 8. Moisture or condensation inside food containers

In our experiment, we understood that, more than the individual fruits dataset and groceries data set, refrigerator storage data has boosted model performance. Refrigerator data alone showed approximately similar results as the total dataset.

### VII. CONCLUSION & IMPACT OF THE MODEL

our system serves as an intelligent grocery tracking system, functioning as a personal assistant to assist residents in their daily food selection process, thereby enhancing their overall quality of life. This proves particularly beneficial for individuals with busy schedules who prioritize their health and prefer home-cooked meals. Moreover, our system offers substantial utility across various customer categories, catering to a diverse range of needs and preferences

### FUTURE SCOPE

We found this field is very vast and there is high scope for improvement and research. Below are some ideas.

1) *Integrate the supply chain groceries prediction system[9] to achieve more accurate predictions of grocery quantities and improve the overall system efficiency.*
2) *Integrate the system with AI based food suggestions system[8] so that based on users' interests in that week, the system can automatically order groceries needed.*
3) *Create an advanced system capable of comprehending complex and cluttered groceries storage closets by utilizing a vast array of high-resolution cameras.*
4) *Create an advanced vision-based system aimed at assessing and verifying the quality and freshness of grocery products.*
5) *By employing motion detection techniques, we can develop an additional system that tracks users' actions and monitors the movement of groceries entering and leaving the closet. The data generated by this system can serve as valuable validation for the predictions*

*made by the main model. Furthermore, this feature can be integrated into an ensemble model, combining it with existing features to enhance overall performance and accuracy.*

ACKNOWLEDGMENT

I would like to acknowledge the support of my mentors who guided me throughout this project research.